\documentclass{article}

\usepackage[nonatbib,preprint]{neurips_2023}

\usepackage[numbers,sort]{natbib}

\usepackage[utf8]{inputenc} %
\usepackage[T1]{fontenc}    %
\usepackage[pagebackref=true,breaklinks=true,colorlinks,bookmarks=false]{hyperref}
\usepackage{url}            %
\usepackage{booktabs}       %
\usepackage{amsfonts}       %
\usepackage{nicefrac}       %
\usepackage{microtype}      %
\usepackage{xcolor}         %

\usepackage{color, colortbl}
\usepackage{graphicx}
\usepackage{soul}
\usepackage{amsmath}
\usepackage{enumitem}
\usepackage{array}
\usepackage{caption}
\usepackage{multirow}
\usepackage{adjustbox}
\usepackage{subcaption}
\newcommand{\PreserveBackslash}[1]{\let\temp=\\#1\let\\=\temp}
\newcolumntype{C}[1]{>{\PreserveBackslash\centering}p{#1}}
\newcolumntype{L}[1]{>{\PreserveBackslash\raggedright}p{#1}}

\newcommand{\audiocc}[1]{{\color{magenta}{}}}

\definecolor{Gray}{gray}{0.9}
\newcommand{\hidecontent}[1]{}
\newcommand{\newremoved}[1]{}

\title{What You Say Is What You Show:\\Visual Narration Detection in Instructional Videos}

\author{%
  Kumar Ashutosh$^{1,2}$ \quad \quad Rohit Girdhar$^{2}$ \quad \quad Lorenzo Torresani$^{2}$ \quad \quad Kristen Grauman$^{1,2}$ \\
  $^{1}$UT Austin, $^{2}$Meta AI\\
}

\begin{document}

\maketitle

\begin{figure}[h!]
\captionsetup{font=footnotesize}
  \centering
  \includegraphics[width=1.0\textwidth]{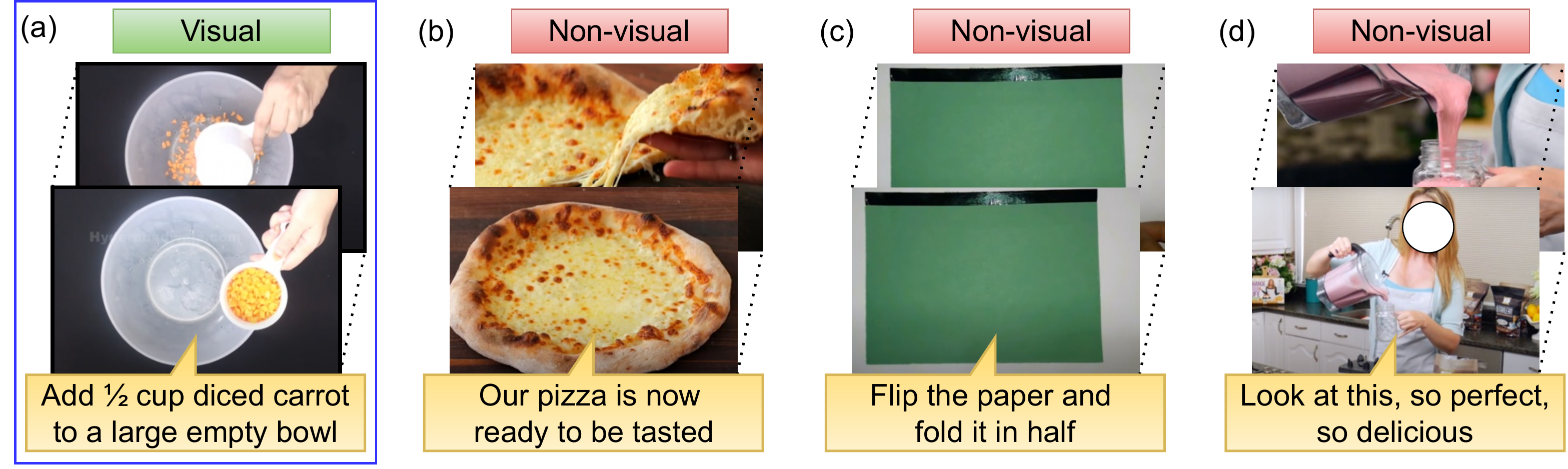}
  \caption{Our goal is to detect \emph{visual narrations}, as in (a), where the spoken narrations describe the action being performed in the video. The other three examples are not visual narrations: (b) the visual and textual signals are relevant to each other, but the action is not visual; (c) the narration states that the paper needs to be folded into half, but it is not demonstrated in the given duration; and (d) the speaker is not talking about her actions. Our \emph{What You Say Is What You Show} (WYS$^2$) model takes a video-narration pair as input and predicts if the narration describes the visible action. %
  }
  \label{fig:examples}
\end{figure}

\begin{abstract}
Narrated ``how-to'' videos have emerged as a promising data source for a wide range of learning problems, from learning visual representations to training robot policies. However, this data is extremely noisy, as the narrations do not always describe the actions demonstrated in the video. To address this problem
we introduce the novel task of {\em visual narration detection}, which entails determining whether a narration is visually depicted by the actions in the video. 
We propose \textit{What You Say is What You Show} (WYS$^2$), a method that leverages multi-modal cues and pseudo-labeling to learn to detect visual narrations with only weakly labeled data. 
Our model successfully detects visual narrations in in-the-wild videos, outperforming strong baselines, and we demonstrate its impact for state-of-the-art summarization and temporal alignment of instructional videos.
\end{abstract}

 \audiocc{We further generalize our approach to operate on only audio input, learning properties of the narrator's voice that hint if they are currently doing what they describe.}
\section{Introduction}

In instructional or ``how-to'' videos, a person demonstrates a procedural activity, such as \textit{making pasta from scratch}, \textit{replacing car tires}, or \textit{using a vacuum cleaner}.  Typically the activity is shown step by step, with helpful narration by the demonstrator, e.g., \emph{``first we add the eggs...now we roll the dough...''}  Today human viewers can learn new skills by watching such videos---a compelling visual alternative to conventional written manuals.  Recent work shows that computer vision models have a lot to learn from instructional videos too.  Being explicit illustrations of activities, they are valuable data sources to train models for activity recognition \cite{yang2020temporal,ghadiyaram2019large,kazakos2019epic}, action anticipation \cite{girdhar2021anticipative,nagarajan2020ego,furnari2020rolling}, and imitation learning \cite{liu2018imitation,torabi2019imitation,nagarajan2021shaping},  while their accompanying verbal narrations offer weak supervision for learning multi-modal representations~\cite{luo2020univl,li2020hero,alayrac2020self} or text-to-video-retrieval models \cite{yang2021taco,zhu2020actbert,miech19endtoend}.

However, as rich as they are, how-to video narrations are not equivalent to activity labels.  
Since instructional videos are designed for human consumption, the narration is structured to maximize both entertainment and skill learning. %
Greeting phrases like \textit{``welcome to my channel!"}, or
\textit{``how y'all doing?"} are frequently present, and the actual demonstration is often interleaved with ancillary comments such as \textit{``this dish smells amazing"} or \textit{``it's my favorite tool for the job"}, which make the narration interesting but do not %
concretely relay the visible task steps. %
Furthermore, the narrator may describe an alternative way to execute a step without demonstrating it, e.g., \textit{``you could also shred the cheese in your food processor"} when instead demonstrating how to manually shred it, or she may refer to non-visual steps like \textit{``next we wait for 30 minutes"}.  In fact, about 50\% of in-the-wild how-to video clips do not show what is being narrated~\cite{miech19howto100m}.  We refer to all such narrations as \emph{non-visual}.

In this context, we introduce a novel task: \emph{visual narration detection}.  The goal is to infer whether a narration describes the actions depicted in a given video segment.  See Figure~\ref{fig:examples}.  Successfully detecting visual narrations would facilitate a number of applications, including automatic video summarization, grounding language in perception, robot learning from demonstration, and prioritizing relevant visual features for activity understanding tasks.  For example, in summarization, one could 
automatically create compact how-to videos that convey only the necessary steps; in robot learning, one could ask a robot to imitate only those steps that are both described and demonstrated.

The proposed task is distinct from learning self-supervised video representations with multiple weakly aligned modalities~\cite{miech19endtoend,alayrac2020self,akbari2021vatt,xu2021videoclip} or computing temporal alignment~\cite{tan}.  Whereas the former work assumes all narrations have a visual counterpart, albeit possibly at some temporal offset~\cite{miech19endtoend},  our task acknowledges that many narrations have no visual instantiation in the video. 
Also note that a narration may be \emph{relevant} or \emph{alignable}~\cite{tan} to the overall task being shown without reflecting a \emph{concrete visual demonstration}---for example, \textit{``make sure you don't wrinkle the paper as you fold'',} or \textit{``our pizza is ready''.}
Further, rather than pretrain a general-purpose video encoder from noisy data~\cite{miech19endtoend,alayrac2020self,akbari2021vatt,xu2021videoclip}, our goal is to learn a model that can identify visual narrations in novel videos.

We propose \textit{What You Say is What You Show} (WYS$^2$), an approach to detect visual narrations.  %
The input is a video segment and its automatically transcribed narration (text), and the goal is to predict whether the video-text pair is a visual narration---that is, \textbf{do the two modalities depict the same action in a task demonstration?}  Though no labeled data exists for our task, we show how to generate training pairs by leveraging a small-scale dataset labeled for demonstration key-steps together with simple language matching to connect them to transcribed narrations.  Then we train a joint video-text embedding that scores the likelihood any new pair constitutes a visual narration.  We further scale up training by using our model to pseudo-label unlabeled instructional videos, allowing our system to harness the power of large-scale uncurated data without human intervention.

We validate our ideas using instructional videos from the CrossTask~\cite{zhukov2019cross}, COIN~\cite{tang2019coin}, and HowTo100M~\cite{miech19howto100m} datasets.
We collect manual annotations on some narrated video data from COIN and CrossTask %
to form a new ground-truth benchmark for visual narration detection, which will be released to the community. Our experiments show that WYS$^2$ consistently outperforms strong baselines on this new task, even when we train it with substantially less data.
Finally, WYS$^2$ also achieves state-of-the-art performance on two downstream tasks: alignability prediction and instructional video summarization, showing the real-world implications of this new task.

\section{Related Work}

\textbf{Learning from instructional video.}
While much work has been devoted to modeling human activity from general Internet videos~\cite{kay2017kinetics,caba2015activitynet,soomro2012ucf101,ghadiyaram2019large}, recent datasets comprised of instructional videos, such as HowTo100M \cite{miech19howto100m}, COIN~\cite{tang2019coin}, and CrossTask~\cite{zhukov2019cross}, have shown the possibility of leveraging this source of data for  
tasks like text-to-video retrieval~\cite{gabeur2020multi,gabeur2022masking,chen2021multimodal,liu2021hit,wang2021dig,bain2021frozen,luo2021clip4clip,miech2021thinking,luo2020univl}, temporal segmentation~\cite{yang2021taco,sun2019learning,luo2020univl,zhu2020actbert,piergiovanni2021unsupervised}, activity localization~\cite{chen2021multimodal,xu2021boundary,zhukov2020learning,yang2021taco,luo2020univl}, anticipation~\cite{girdhar2021anticipative,sener2019zero,wu2022memvit,dessalene2021forecasting,furnari2020rolling}, question-answering~\cite{yang2021just,seo2021look,li2020hero}, and summarization~\cite{iv-sum,clip-it}.
None of the prior work explores visual narration detection, as we propose.  Our idea could potentially augment any of these existing instructional video models by focusing their attention on video segments that illustrate the narration, as we demonstrate concretely for %
video summarization \cite{iv-sum}.

{\bf \noindent Associating vision and language.}  A wide body of literature explores creative ways to link language and vision. %
Whether attaching words and phrases to 2D images~\cite{luo2020multi,referring-expressions-murphy,yu2018mattnet,shi2018key,wang2019neighbourhood,dai2017detecting,dong2021visual} or captioning images~\cite{anderson2018bottom,huang2019attention,yao2018exploring,li2019entangled,zhou2020more} and videos~\cite{seo2022end,luo2020univl,huang2020multimodal,korbar2020video,tang2021decembert,hessel2019case}, a fundamental goal is to discover associations between how things look and how they are described.  Beyond descriptions, question answering brings common sense together with vision~\cite{lei2018tvqa,lei2019tvqa+,lu2016hierarchical,Antol_2015_ICCV,yu2019deep,das2018embodied,das2018neural,yu2019multi}, while agents performing vision and language navigation translate instructions into movement~\cite{anderson-vln-1,anderson-vln-2,anderson-vln-3,are-you-looking}. In grounded language learning, language models are acquired in the context of perceived objects, events, and environments~\cite{tellex,kebe2021a}. 
Visual narration detection can be seen as a  new task for relating vision (human demonstrations in video) to language (spoken narrations of action).  

{\noindent \bf Multi-modal video embeddings.}
Multi-modal data accompanying video is ripe for learning enriched visual representations.  Self-supervised models leverage audio and vision to pretrain video encoders~\cite{morgado-spatial-nips2020,lorenzo-nips2020,korbar-nips2018}. Text-video(-audio) embeddings~\cite{mithun2018learning,miech2018learning,alayrac2020self,hiervl,egovlp} have also been explored, with implications for downstream tasks in text-to-video retrieval or captioning~\cite{miech19endtoend,miech19howto100m,alayrac2020self,rouditchenko2020avlnet,xu2021videoclip,patrick2020support,luo2020univl,gabeur2020multi}.
While uncurated instructional video is appealing for training multi-modal embeddings, a technical challenge is that the modality streams need not be exactly aligned temporally.  For example, a person might describe a step and then demonstrate it seconds later.  Recent work~\cite{alayrac2020self,akbari2021vatt,sun2019learning} including MIL-NCE~\cite{miech19endtoend} and VideoCLIP~\cite{xu2021videoclip} address the cross-modal misalignment head-on, either using multiple instance learning to represent the alignment ambiguity~\cite{miech19endtoend,alayrac2020self,akbari2021vatt,luo2020univl}  
or sampling longer video windows around a narration to find the best match~\cite{xu2021videoclip}.
Their goal is to pretrain video encoders, and none address visual narration detection; in fact, they implicitly assume that every narration has a visual counterpart. %
Relaxing that assumption, temporal alignment networks (TAN)~\cite{tan} distinguish between ``alignable" and non-alignable text, but this is a different and looser notion of agreement than visual narrations.
For example, \emph{``look at these fresh tomatoes''} may be \emph{aligned} with video but does not depict a \emph{demonstrated action}, e.g., 
will not help a user learning to make tomato soup.

{\noindent \bf Predicting visualness.}
While no prior work addresses visual narration detection, other notions of ``visualness'' have been explored.  %
Given a caption, plus or minus an image, the \emph{visual text} model~\cite{dodge2012detecting} infers if a word appears concretely in the image. For example, \textit{car} is visual in an image captioned \textit{``here's my dream car''} but not so in \textit{``view from my car's window''}.   The sports commentary system of~\cite{chen2008learning} learns language from sportscasts of simulated soccer games, discovering association between commentary and game states, and filtering out \emph{non-visual} commentary like \textit{``player1 looks around for a teammate''}.  The VQA system of~\cite{ray2016question} identifies non-visual questions (e.g., \textit{``who is the president of this country?''}) in order to decide whether the image content should be relevant for answering the question.  We introduce a new form of visualness---relating human demonstrations to narrations---which demands video understanding on in-the-wild instructional video clips.
\section{Technical Approach}

We propose \textit{What You Say is What You Show} (WYS$^2$), a method to detect visual narration in instructional videos. As discussed above, there are many instances %
where the narration digresses from what is being shown, whether to keep the listener engaged, provide alternative how-to advice, or simply describe things that cannot be physically demonstrated in the video (see Figure~\ref{fig:examples}).

Our objective is to identify such video segments using as input the video-text pair \audiocc{(or alternatively, the narration audio alone, as we discuss in Sec.~\ref{sec:embedding})}.  Formally, let $v$ denote a video clip, $a$ the audio from the same temporal segment, and $t$ the text resulting from automatic speech recognition (ASR) applied to $a$.
We aim to learn a visual narration detector,  $\mathcal{V}_{vt} : (v, t) \rightarrow \{0, 1\}$, where $\mathcal{V}_{vt}(v, t) = 1$ when the narration $t$ describes the activity demonstrated in the video $v$, and $\mathcal{V}_{vt}(v, t)=0$ otherwise (i.e., \emph{is the person describing what they are showing?}).  

In the following, we first show how to obtain suitable training data for this new task through a combination of bootstrapping a small partially labeled dataset together with pseudo-labeling on a large-scale unlabeled collection of videos (Sec.~\ref{sec:data}).  Then we formulate WYS$^2$ as a multimodal embedding learning problem %
(Sec.~\ref{sec:embedding}).

\subsection{Visual Narration Training Data}
\label{sec:data}

\begin{figure}[t]
\captionsetup{font=footnotesize}
    \centering
    \begin{minipage}{0.4\textwidth}
\centering
    \begin{tabular}{cc}
\includegraphics[width=0.45\textwidth]{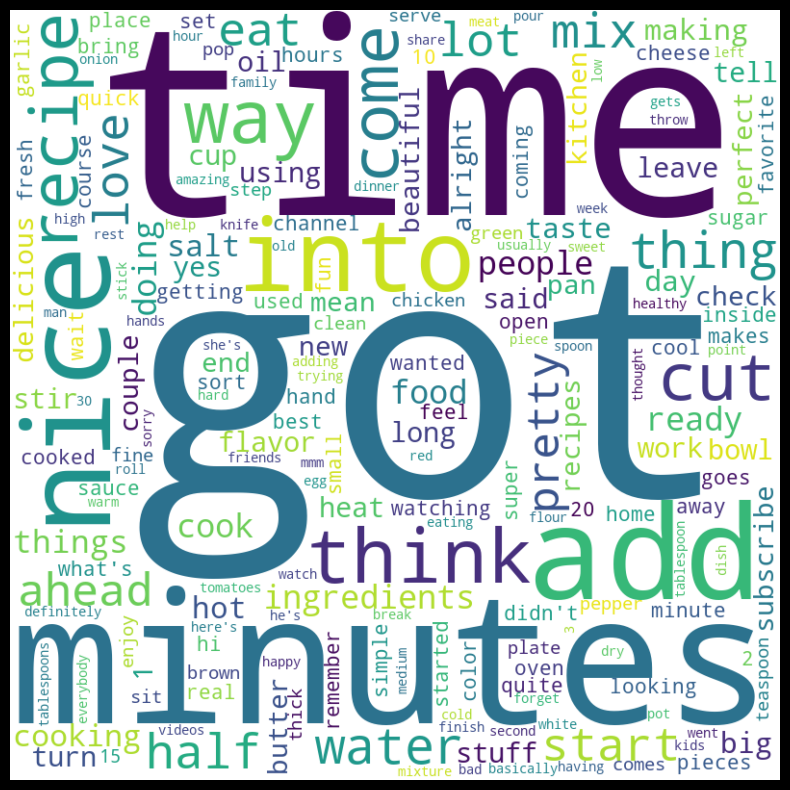}

\includegraphics[width=0.45\textwidth]{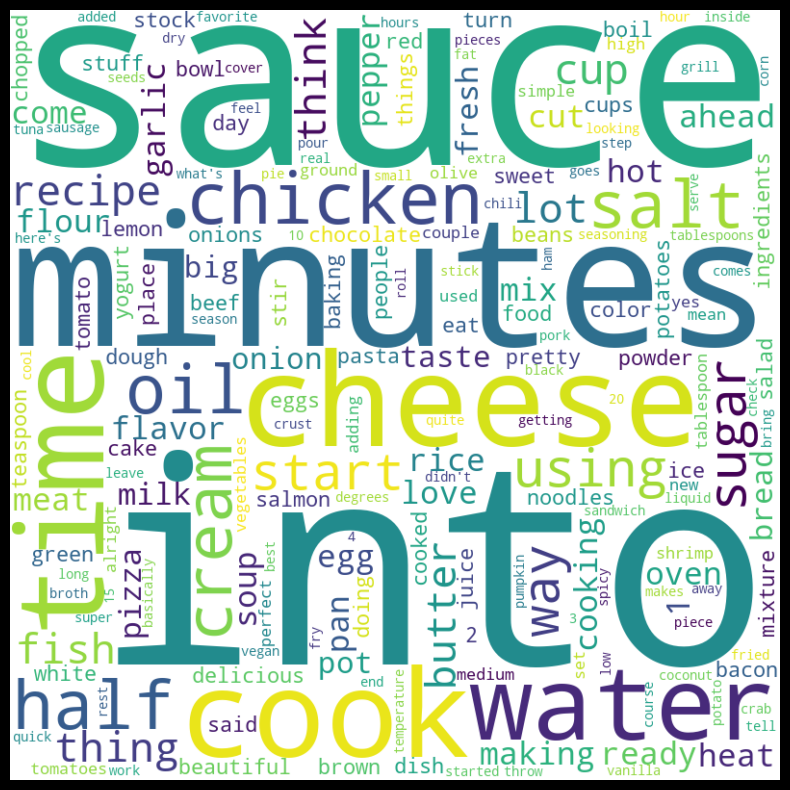}
\end{tabular}
\caption{Words in non-visual (left) and  visual (right) narrations.  %
} %
\label{fig:word-cloud}
    \end{minipage}\hfill
    \begin{minipage}{0.58\textwidth}
        \centering
        \includegraphics[width=0.9\textwidth]{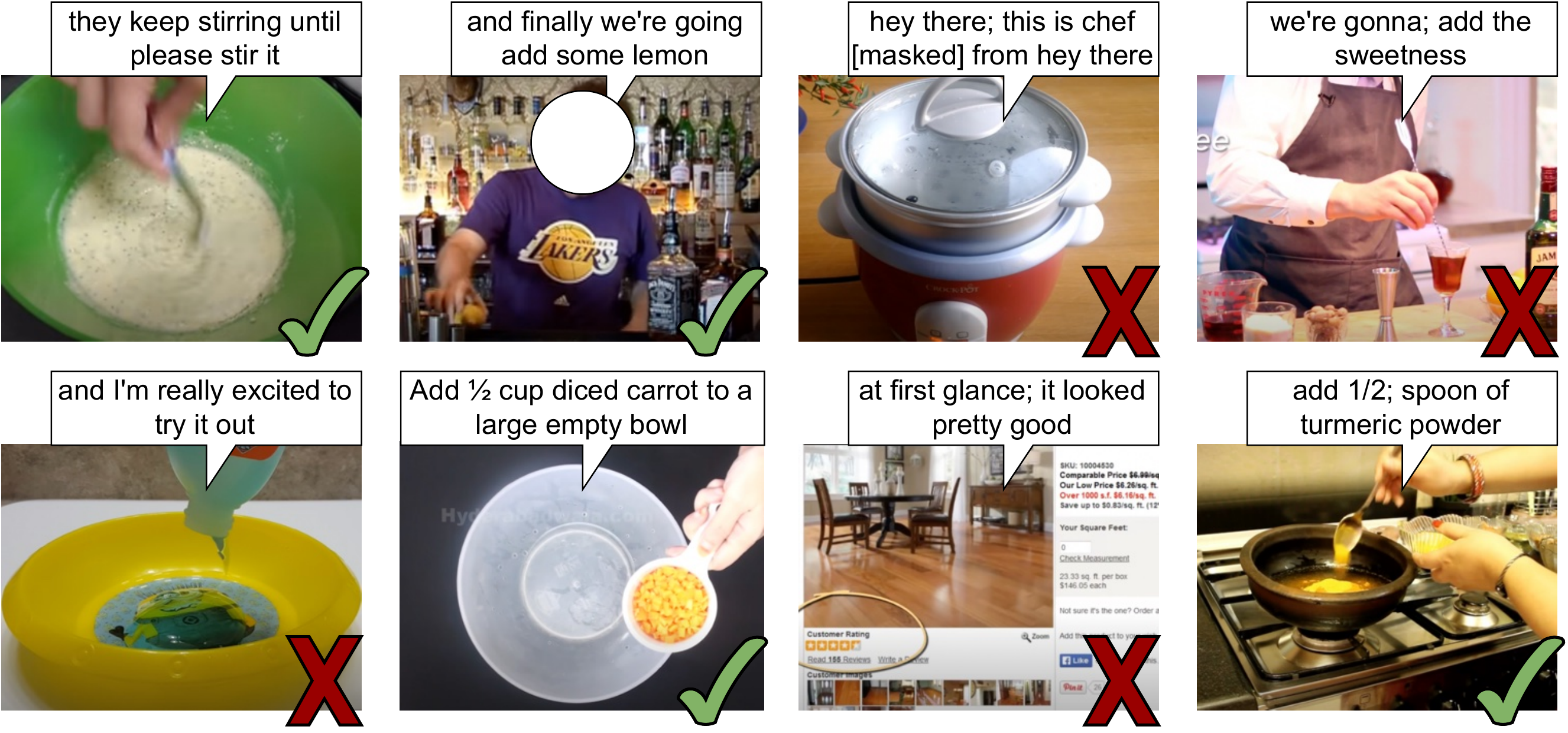} %
        \caption{Example clips from manually-annotated gold standard VND test set.}
        \label{fig:mturk-examples}
    \end{minipage}
\vspace{-0.4cm}
\end{figure}

There are no current datasets that can be directly used to learn  $\mathcal{V}$. Thus, we propose a method that automatically infers visual narration labels from existing %
datasets annotated with demonstration key-steps. We show that a model trained on these derived visual narration labels can then be applied to \emph{pseudo-label} large-scale uncurated data.  We further boost the accuracy of our visual narration detector by retraining it on the aggregation of the smaller labeled dataset and the  pseudo-labeled large-scale data.  We stress that while our method learns to recognize visual narration {\em without} requiring manual labeling by leveraging existing key-step annotations and pseudo-labels, in our experiments we validate our predictions against manually-collected ground-truth annotations (Sec.~\ref{expts} and Fig.~\ref{fig:mturk-examples})  which will be released to the
community.

\begin{figure*}[t]
\captionsetup{font=footnotesize}
\centering
\includegraphics[width=\textwidth]{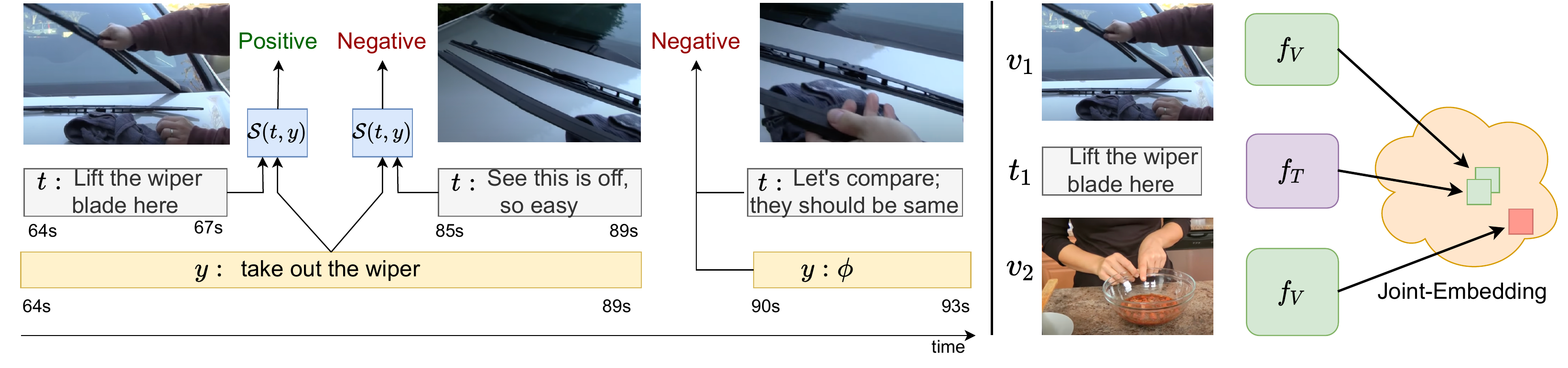}
\vspace*{-0.25in}
\caption{Schematic representation of our WYS$^2$-SS approach. For every pair of narration and key-step $(t, y)$ we compute $\mathcal{S}(t, y)$ and choose positive clips (left). Similar video and text segments are close in the joint-embedding space (right).} %
\label{fig:method}
\vspace{-0.15in}
\end{figure*}

\textbf{Visual Narration Labels from Key-Steps.} Let $\mathcal{D}_L = \{ (v^{(i)}, t^{(i)}, y^{(i)}, o^{(i)}) \}_{i=1}^{|\mathcal{D}_L|}$
be a dataset where along with video clip $v^{(i)}$ and the spoken narration $t^{(i)}$, we have manually supplied textual descriptions $y^{(i)}$ of the action demonstrated in $v^{(i)}$, a.k.a. annotated key-steps, as well as task-level labels $o^{(i)}$. Both $o$ and $y$ come from a respective predefined taxonomy.  For example, for a clip $v$ in Figure~\ref{fig:method} the first narration $t$ is \emph{``lift the wiper here..."} and the key-step annotation $y$ is \emph{take out the wiper}.  The coarser task-level label $o^{(i)}$ \emph{how to change wiper blades} applies to that clip and all those in the remainder of the video.  Note that only a subset of clips have a key-step label, and when a key-step is labeled, it means the labeled action appears in the camera's field of view~\cite{tang2019coin,zhukov2019cross}.
We explore three variants of WYS$^2$, which differ in how they formulate the inferred visual narration labels for training. The second and third variants are ablations of the first variant.

The first variant, WYS$^2$-SS, infers correspondences between key-step text and narration text.
Note that $y^{(i)} = \phi$ when no key-step is present in the video clip $v^{(i)}$, while a non-null $y^{(i)}$ implies the video clip $v^{(i)}$ contains a visual demonstration of a key-step. However, as shown in Figure \ref{fig:examples}(c), although the video may demonstrate an activity $y$, the narration $t$ may fail to describe it. Such a case should be recognized as a ``negative'' sample for visual narration detection. Conversely, a ``positive'' visual narration sample %
occurs when the narration matches the annotated key-step, i.e., when $t^{(i)}$ and $y^{(i)}$ convey the same thing. 

Since the key-step language and narration will not be identical lexically, we  assess their semantic agreement to discover matches.
Specifically, 
we leverage a sentence similarity function $\mathcal{S}: (t, y) \rightarrow [0,1]$. This provides us with a mechanism to automatically derive a training set $\mathcal{D}_{vn}$ of positive visual narration samples from the initial dataset $\mathcal{D}_L$ annotated with key-steps:
$$ \mathcal{D}_{vn}^{SS} = \left\{ (v, t)~~\vline~~ (v, t, y) \in \mathcal{D}_L ~~\text{and}~~ \mathcal{S}(t, y) \geq c   \right\}, $$
where $c$ is a hyperparameter controlling the amount of samples declared to be positive, and superscript $SS$ denotes ``sentence similarity". 
To generate negative samples, we pair video and text from different clips in the same batch to speed up the training process, following the standard practice~\cite{miech19endtoend,luo2020univl,sun2019learning}.
For the sentence similarity function $\mathcal{S}$, %
we first resolve coreferences~\cite{Shen-VisualNarrationProceL:CVPR21,doughty2020action} then use a pre-trained model from InferSent \cite{conneau-EtAl:2017:EMNLP2017}, an English sentence embedding model.
The process is shown in Figure \ref{fig:method}.

The second variant, WYS$^2$-VR, is an ablation of WYS$^2$-SS that does not account for the similarity between $y$ and $t$.  Here %
we consider all training clips having a key-step annotation as a positive sample:
$$ \mathcal{D}_{vn}^{VR} = \left\{ (v, t) ~~\vline~~ (v, t, y) \in \mathcal{D}_L ~~\text{and}~~ y \neq \phi \right\}.$$
In other words, the positive samples are all those known to be \emph{visually relevant} (VR) to the task being demonstrated, since they belong to one of its key-steps---a lower bar.

The third variant, WYS$^2$-MC, infers visual narration labels using the consensus of per-modality task-level classifiers, 
where MC denotes ``modality consensus.''    Recall that $o^{(i)}$ indicates the overall skill%
---for example, \textit{how to change wiper blades} or \textit{how to make banana ice cream}. %
WYS$^2$-MC trains 
a classifier for each modality: video clip classifier $f_C^{V}(v; \theta^V)$ %
and sentence classifier $f_C^{T}(t; \theta^T)$, both of which target the task labels.  %
The top high scoring video-text pairs $(v, t)$ are deemed positive: %
\begin{align*}
    \mathcal{D}_{vn}^{MC} = \{ (v, t) ~~\vline~~ (v, t, o) \in \mathcal{D}_{L} ~~&\text{and}~~\textrm{rank}(f_C^{V}(v; \theta^V)) \leq k %
    ~~\text{and}~~ \textrm{rank}(f_C^{T}(t; \theta^T)) \leq k \}.
\end{align*}%
In our experiments, we use SlowFast \cite{fan2020pyslowfast} for the multi-class video classifier $f_C^V$ and Wikipedia2Vec \cite{yamada2019neural} for the multi-class sentence classifier $f_C^T$. We also experimented with word2Vec \cite{mikolov2013efficient} and InferSent \cite{conneau-EtAl:2017:EMNLP2017} but found Wikipedia2Vec yields superior performance. %
Hyperparameter $k$ is set using validation data. %

Figure~\ref{fig:word-cloud} shows narration words from sentences classified as least (left) and most (right) visual for the datasets introduced in Sec.~\ref{expts}.
Descriptive terms like \textit{nice}, \textit{love}, \textit{pretty}, \textit{time} occur commonly in low-ranked narrations while action-specific terms like \textit{sauce}, \textit{cheese}, \textit{chicken}, \textit{cook} are often associated with high-ranked narrations. %
As we will see in results, however, words alone are insufficient to predict if the narrated actions are visually depicted.

We stress that the input key-step labels are \emph{not} the same as visual narration labels; part of our contribution is to bootstrap them for our task by the process above.   In Sec.~\ref{expts} we show that these simple procedures applied to existing small-scale datasets with annotated key-steps enable learning an effective model for visual narration detection. Below, we  describe how to further boost the accuracy of the detector by means of pseudo-labeling. %

\textbf{Visual Narration Pseudo-Labels on Unlabeled Videos.} Let $\mathcal{V}$ denote a preliminary visual narration detector trained on $\mathcal{D}_{vn}$ (the exact form of $\mathcal{V}$ and training objective will be given in Sec.~\ref{sec:embedding}).  
To scale up training, we next leverage a large-scale {\em unlabeled} instructional video dataset $\mathcal{D}_U = \{ (v^{(i)}, t^{(i)}) \}_{i=1}^{|\mathcal{D}_U|}$, together with $\mathcal{D}_{vn}$ and detector $\mathcal{V}$.
We use $\mathcal{V}$ to generate visual narration (pseudo)-labels for the whole dataset. We then append the positive video clips into a larger training dataset $\mathcal{D}^\prime_{vn}$: %
$$\mathcal{D}^\prime_{vn} = \mathcal{D}_{vn} \cup \left\{ (v, t) ~~\vline~~ (v, t) \in \mathcal{D}_U ~~\text{and}~~ \mathcal{V}(v, t)=1  \right\}.$$
Using the predictions as pseudo-labels dramatically increases our training data, allowing us to harness the power of large-scale unlabeled datasets without human intervention.  Doing so significantly boosts performance, as we will show in the results section.

\subsection{Multimodal Visual Narration Detector}
\label{sec:embedding}

Next, we present the model and training objective for $\mathcal{V}$. In our ablation experiments, we pair the training methodology below with each of the three variants for inferring labels presented above.

\audiocc{\textbf{Inferring Visual Narrations from Video-Text Pairs.}} Our visual narration detector $\mathcal{V}_{vt}$ consists of two subnetworks---one for extracting the video embedding $f_V(v)$ and the other for obtaining a sentence embedding $f_T(t)$.  The idea is to learn $f_V$ and $f_T$ such that clips with a visual narration have high similarity in the embedding space, while the others have low similarity.  See Figure~\ref{fig:method}(right). %

We use the S3D \cite{xie2018rethinking} architecture to compute the visual embedding from video clips. To obtain text embeddings, we use a pre-trained Google News word2vec model \cite{mikolov2013efficient}.  
 The word2vec embedding is passed through a linear layer (with ReLU activation) and then max-pooled to obtain the sentence embedding. The video architecture is learned while the sentence encoder is frozen. %
 The choice of these embedding architectures is consistent with recent work \cite{miech19endtoend,xu2021videoclip,luo2020univl,alayrac2020self,gabeur2020multi,bain2021frozen} and suitable for our objective of multimodal embedding learning. 
Following recent success in contrastive learning, we use softmax-type Noise-Contrastive Estimation (NCE) \cite{oord2018representation} as our loss function:
$$-\sum_{i=1}^{n}\log\left( \frac{e^{f_V(v^{(i)})^Tf_T(t^{(i)})}}{e^{f_V(v^{(i)})^Tf_T(t^{(i)})} + \sum_{(v, t) \in \mathcal{N}^{(i)}}e^{f_V(v)^Tf_T(t)}} \right),$$
where $\mathcal{N}^{(i)}$ is the sample-specific negative set chosen from the same batch.   
 During inference, we compute the similarity score using a dot product, i.e., $f_V(v)^Tf_T(t)$ and apply a threshold to detect instances of visual narration.

\audiocc{\textbf{Inferring Visual Narrations from Audio.} We separately consider an alternative setting where the input to the model is solely the audio waveform.
Here the goal is to predict whether the (unobserved) video is likely to show what is being described
(i.e., \emph{does the audio suggest that the person is showing what they are describing}?).
We hypothesize that audio elements such as vocal intonation, cadence, and pacing may signal when the narrator's speech is demonstration-relevant versus merely color commentary.  For example, the narrator may slow down or talk more deliberately when declaring a key step they are showing.  Similarly, 
an instruction has a different tone than welcoming and thanking the viewers.  
This setting is attractive for low-power scenarios, since audio streaming is much more efficient than for video. This variant could enable applications that stream only audio for the non-visual parts, thereby saving on storage and transmission requirements.  %

Formally, for the audio-input case, we aim to learn an audio-based visual narration detector $\mathcal{V}_a : a \rightarrow \{0,1\}$, where $a$ denotes the raw audio waveform from the narration, and the target labels are  determined by the video-text agreement, as before. 
To train $\mathcal{V}_a$, we first extract MEL-Spectrograms from the audio waveform $a$ and pass the 2D array to a standard ResNet-18 \cite{resnet} architecture. To compute the binary classification output we attach a fully-connected layer with sigmoid activation function. The standard Binary Cross Entropy (BCE) loss is used to train the model to match the visual narration labels in $\mathcal{D}_{vn}$. 

Our insight differs from the more common observation that audio can enhance action recognition~\cite{kazakos2019epic,nagrani2021attention,gao2020listen}
(e.g., hearing a pan sizzling reinforces the recognition of the action \emph{stir-frying} from the visual stream).  Instead, we aim to capitalize on the narration properties that signal \emph{both} the video and text content are aligned.  Hence, it is significant that our input is \emph{audio} for this variant, rather than the narration's transcribed text---the latter would not reveal these category-agnostic aspects of how the speech is delivered.
}

\section{Experiments}
\label{expts}

\label{expsetup}

\noindent\textbf{Datasets.}
\textbf{COIN} \cite{tang2019coin} and \textbf{CrossTask} \cite{zhukov2019cross} are %
instructional video datasets containing 11,827 and 2,750 videos, respectively.\footnote{Since some were deleted or made private by the owners, we use all currently available videos: 10,353 for COIN and 2,546 for CrossTask.} The videos contain an array of daily life how-tos, such as cooking, repairs, personal hygiene, and sports. 
They have key-step annotations $y$ drawn from taxonomies containing a total of 883 unique key-step types, i.e., temporal windows where one step of the demonstration occurs, such as \textit{wipe off dipstick}, \textit{put bananas into blender}, \textit{lift the wiper}, etc.
We aggregate these two datasets to form the labeled dataset $\mathcal{D}_L$. This yields a collection with a total of 415 hours of video,  146 hours of which are annotated with key-step descriptions. %
By applying the steps in Sec.~\ref{sec:data}, we derive our visual narration dataset $\mathcal{D}_{vn}$, which totals 32 hours and 115k clips (SS), 117 hours and 420k clips (VR), and 150 hours and 540k clips (MC).  %
We first take out $20\%$ of the dataset for creating a gold standard VND Test Set (explained next) and  we split the rest into training and validation sets in a 80:20 ratio, denoted  $\mathcal{D}_{vn}^{tr}, \mathcal{D}_{vn}^{val}$. %
\textbf{HowTo100M} \cite{miech19howto100m} is a collection of 1.22M videos totaling 136M video clips. %
It also contains videos from daily life how-tos, but %
does not contain any temporal annotation. 
We use this data as our unlabeled dataset $\mathcal{D}_U$ for pseudo-labeling (cf.~Sec.~\ref{sec:pseudo}).

\newremoved{
\begin{figure*}[t]
  \centering
  \begin{subfigure}{0.45\linewidth}
    \includegraphics[width=0.95\linewidth]{figures/fig-4a.pdf}
    \caption{Comparison to baselines}
    \label{fig:short-a}
  \end{subfigure}
  \hfill
  \begin{subfigure}{0.45\linewidth}
    \includegraphics[width=0.95\textwidth]{figures/fig-4b.pdf}
    \caption{Ablation}
    \label{fig:short-b}
  \end{subfigure}
  \caption{\cc{Visual narration detection accuracy. (a) Our WYS$^2$ model outperforms baselines trained using multimodal embedding objectives from the literature for temporal alignment~\cite{miech19endtoend} and overlapping clips~\cite{xu2021videoclip}.   (b) \KGnew{Ablation showing that} the WYS$^2$-SS variant performs best, showing the effectiveness of our approach to link the key-steps and spoken narrations to seed training.} \KAnote{We can remove the figure and have table instead since we have newer results} }
  \label{fig:output-pr-curve}
\end{figure*}
}
\noindent\textbf{Gold Standard VND Test Set.}
One of our contributions is to automatically generate labels to train a visual narration detection (VND) model using minimal manual annotations for a task, i.e., key-step labels on a smaller-scale collection (Sec.~\ref{sec:data}). To rigorously evaluate our model, however, it is important to have a gold standard test set with visual narration labels determined by human observers.  To that end, we collected a gold standard VND test set %
via crowdsourcing using the 20\% withheld COIN and CrossTask data noted above.
We ask seven MTurkers whether each video segment has the exact demonstration of the actions in the accompanying narration. 
We require a consensus of at least five out of seven annotators to assign a positive or negative label to a clip.
This process yields  5.4 hours of test data consisting of 3,352 positive and 2,916 negative manually labeled video-text test pairs. We find that 76\% of WYS$^2$-SS model's automatically inferred positive labels are consistent with the %
manually labeled pairs, reinforcing %
our model's reliability for training. See Fig. \ref{fig:mturk-examples} for examples and Supp. for data-collection details. %

\noindent\textbf{HTM-Align Test Set.} TAN \cite{tan} provides a manually annotated \emph{alignment} dataset  called HTM-Align composed of 80 videos from HowTo100M~\cite{miech19howto100m}. %
We also show results on 
HTM-Align for both visual narration detection and the authors' proposed task of alignability prediction.

\paragraph{Baselines.}
Since no prior work performs visual narration detection, as baselines we consider alternative approaches for deriving training sets for VND from $\mathcal{D}_L$, using ideas from the literature originally explored for representation learning~\cite{miech19endtoend,alayrac2020self,akbari2021vatt,luo2020univl,xu2021videoclip} and video alignment~\cite{tan}. We also compare with a text-only baseline (GPT-2 \cite{gpt2}) and a visual-only object detection baseline \cite{detic}. %

\begin{itemize}[leftmargin=*]
    \item     \textbf{Temporal Alignment (MIL-NCE):} Used in \cite{miech19endtoend,alayrac2020self,akbari2021vatt,luo2020univl}, this method assumes that {\em each} video clip has a corresponding narration that describes it, albeit potentially misaligned in time. To handle misalignments, a multiple instance learning (MIL) variant of the NCE loss is proposed \cite{miech19endtoend}.
We set the number of candidate positives $|\mathcal{P}_i| = 3$ based on the validation set.\vspace*{-0.07in}

\item \textbf{VideoCLIP Overlapping Clips (OC):} 
Proposed in~\cite{xu2021videoclip}, this VideoCLIP method takes a narration $t^{(i)}$ centered at frame $i$, extracts a long video clip around it, and labels it as a positive sample. The idea is that misalignment is generally limited to a few seconds, and that is alleviated by taking a longer overlapping  clip. %
Following~\cite{xu2021videoclip}, this baseline uses an NCE training objective and 16s clips.\vspace*{-0.07in}
\item \textbf{TAN~\cite{tan}:} A new method to infer temporal alignment between video and narrations while allowing for ``unalignable'' segments.\vspace*{-0.07in}

\item \textbf{CLIP~\cite{clip}:} A popular large-scale video-language embedding model.\vspace*{-0.07in}
\item \textbf{Language-only GPT-2}~\cite{gpt2}: This baseline tests whether narrations alone are sufficient to detect visual narrations (without looking at the video).  We apply a pretrained language embedding to the narration text, then  
train a linear classifier head using the same labels as our model. \vspace*{-0.07in}

\item \textbf{Object Detector}:  Using the Detic visual object detector~\cite{detic}, this baseline classifies a
video segment as positive if at least 60\% of objects mentioned in the narration are also visually present. %
\vspace*{-0.07in}
\end{itemize}

\noindent \textbf{Implementation Details.} Unless stated otherwise, all models are trained with a batch size of 128 on eight Quadro RTX GPU (23 GB each) with an Adam optimizer \cite{kingma2014adam} and learning rate $10^{-5}$. Each training clip has a duration of 3s and $c=0.5$ for the sentence similarity $\mathcal{S}$. 
We generate 4 negative clips per positive based on validation performance.
For training $\mathcal{V}_a$, we use a batch size of 16 and use one GPU with SGD optimizer, a learning rate of $10^{-2}$ and dropout $0.1$. %

\noindent \textbf{Evaluation Metric.} We use ROC-AUC as the evaluation metric. Each model predicts a score and the video segment is labeled ``visual narration'' if it exceeds a threshold. ROC-AUC is a threshold-independent metric summarizing results over all thresholds.

\begin{figure*}[t]
\captionsetup{font=footnotesize}
\centering
\includegraphics[width=\textwidth]{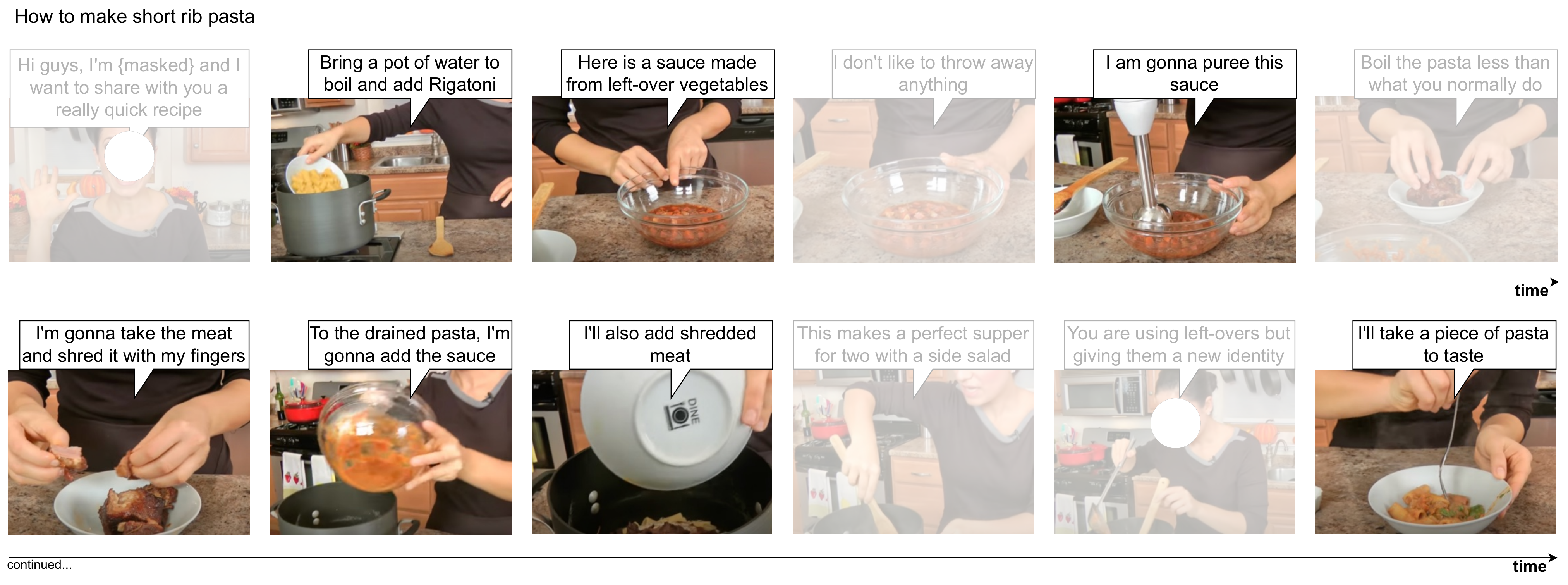}
\includegraphics[width=\textwidth]{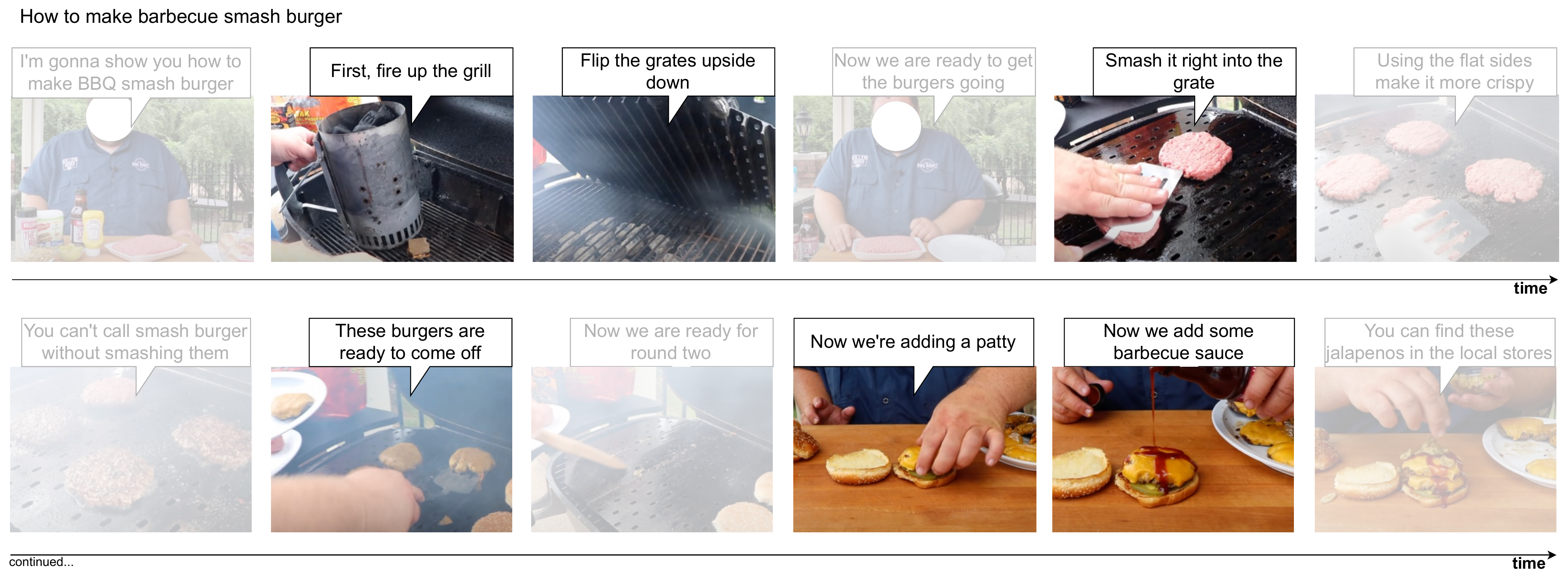}
\includegraphics[width=\textwidth]{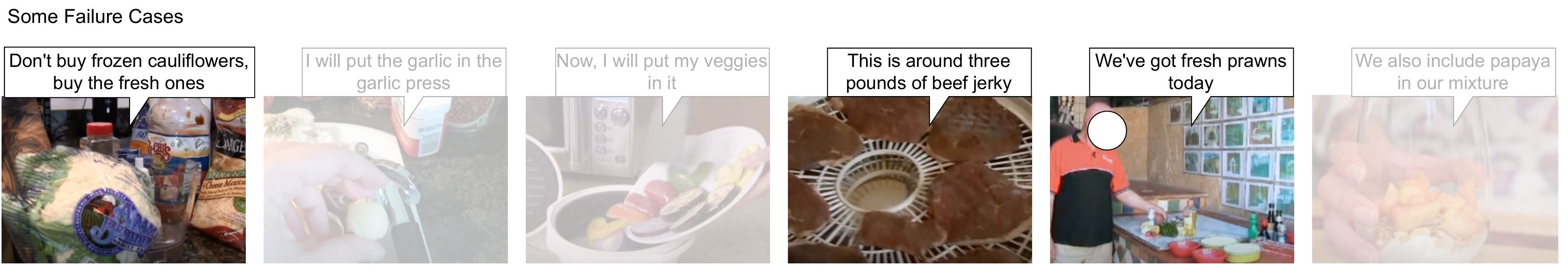}
\vspace*{-0.25in}
\caption{Example of labelling unstructured HowTo100M videos for \textit{visual narrations} for making short rib pasta (top) or smash burgers (middle) %
followed by a row of \textbf{failure case} keyframes from various videos. 
Samples predicted as negatives by our model are shown with transparency. Our model can automatically highlight key-steps in unseen videos, thus motivating a summarization application (cf.~Table~\ref{tab:summarization}). Best viewed with zoom.  More in Supp. } \vspace*{-0.15in}%
\label{fig:outputs}
\end{figure*}

\subsection{Visual narration detection}

Table~\ref{tab:visual-baseline} shows the visual narration detection results.
WYS$^2$-SS outperforms the baselines.%
We attribute our advantage over Temporal Alignment (which adopts the MIL-NCE objective~\cite{miech19endtoend}) 
to the fact that many training clips do not have a visual narration, leading to false positives that cannot be overcome by the alignment tolerance of MIL. %
Similarly, compared to VideoCLIP-Overlapping Clips (which adopts the video padding idea of~\cite{xu2021videoclip}), our model more accurately learns visual narration associations because it does not assume a positive must always exist. %
Moreover, the language-only and object detection baselines %
perform much worse, indicating that the problem requires multi-modal understanding
and joint learning between the modalities, respectively.%
\footnote{Unlike all the methods tested in Table~\ref{tab:visual-baseline}, the TAN \cite{tan} and CLIP \cite{clip} baselines
rely on large-scale pretraining; hence we reserve comparisons to those models for the next section,
where we train methods on all HowTo100M videos for an apples-to-apples comparison.}

An ablation of the variants we explored for generating visual narration labels revealed that WYS$^2$-SS, which relies on sentence similarity to align narrations and key-step labels, performs best (62.7 ROC-AUC on VND Test for SS, compared to 58.8 for VR and 59.3 for MC).
Even though WYS$^2$-VR uses visual annotation, it uses less supervision than WYS$^2$-SS because the latter also exploits ASR sentences for visual narration detection. WYS$^2$-MC does well, but (unlike WYS$^2$-SS) has the disadvantage of being task-specific due to its reliance on task labels $o$.
See Supp.~for further ablations on the selectivity for positive/negative clips, batch size, learning rate. In all further results we deploy WYS$^2$-SS since it performs the best. %

Figure~\ref{fig:outputs} shows examples using our model to detect visual narrations in unseen HowTo100M videos.  We see that WYS$^2$ automatically highlights the clips where the narrations are acted out.  This visualization suggests natural downstream applications of our idea for storyboarding and video summarization, which we will quantitatively show in Sec.~\ref{sec:summarization-results}.   The fact that our model is category-agnostic---able to generalize to new tasks without knowing their key-steps---is essential to this capability.  For example, \textit{making barbecue smash burger} is not present in CrossTask or COIN, yet our model generalizes to detect visual narrations for that task in HowTo100M. 
Figure \ref{fig:outputs}(bottom) shows \textbf{failure cases.} %
Since we use sentence similarity to generate labels for our training, %
sentences that tend to contain words similar to those in key-steps can be mistakenly marked as positive, e.g. \textit{``we've got prawns, three pounds of beef jerky''}.

Overall, these results are encouraging, given the difficulty of the task, which goes beyond traditional activity recognition. While a human observer has no trouble discerning when the instructor is describing their actions versus when they are not, the distinction can be quite subtle.  Often the non-visual commentary will still refer to task-relevant objects or actions (like the \emph{prawns} above), and camerawork may intermittently obscure the described steps.
Furthermore, we stress that visual narration detection is category-agnostic: the same model must predict visualness no matter the task being performed by the demonstrator, as opposed to learning task-specific cues.

\audiocc{

}

\begin{table}[t]\footnotesize
  \captionsetup{font=footnotesize}
    \begin{minipage}{.39\textwidth}
      \centering
      \caption{Visual narration detection on VND Test (ROC-AUC). Our WYS$^2$ model outperforms %
    multimodal embedding objectives from the literature~ \cite{miech19endtoend,xu2021videoclip} as well as baselines considering object agreement in the narrations and video~\cite{detic} or the narration text alone~\cite{bert}. %
    }
    \label{tab:visual-baseline}
    \begin{tabular}{L{3.2cm} C{1.4cm}}
    \toprule
Method &  \footnotesize{ROC-AUC} \\
\hline
  Obj. Detector \cite{detic} & 51.0\\ 
  GPT-2 \cite{gpt2} & 53.5\\ 
  Temporal Alignment \cite{miech19endtoend} & 59.6  \\
  VideoCLIP OC~\cite{xu2021videoclip} & 59.0 \\
  \rowcolor{Gray}
  WYS$^2$-SS (Ours) & \textbf{62.7} \tiny{$\pm$~0.3}\\
 \hline
\end{tabular}
\vspace{0.35cm}
    \end{minipage}
    \hfill
    \begin{minipage}{.60\textwidth}
      \centering
      \caption{Visual narration detection and alignability prediction ROC-AUC. For visual narration detection, we evaluate fine-tuned results 
    with pseudo-labels on the VND (left column) and HTM-Align (middle column) test sets. We outperform baselines on both the test datasets. Right column shows the alignability prediction following the autors' setting \cite{tan}. Our model outperforms all prior methods. %
    }
    \label{tab:vnd-and-align}
\begin{tabular}{L{2.4cm}  C{1.1cm}  C{1.1cm}  C{1.1cm}}
\toprule
& \multicolumn{2}{c}{\small{VND}} & \small{Align.} \\
 \cmidrule(r){2-3} \cmidrule(r){4-4}
&  \small{VND-T %
}& \small{HTM-A %
} & \small{HTM-A}\\
\midrule
Obj. Detector \cite{detic} &52.5&52.5&52.8\\
GPT-2 \cite{gpt2} &63.2&70.7&65.9\\
  CLIP \cite{clip}& 66.8 & 76.0 & 71.7\\ 
  MIL-NCE \cite{miech19endtoend} & 73.6 & 79.5 & 73.1  \\ 
  TAN \cite{tan} & 69.7 & 77.4 & 75.1\\
  \rowcolor{Gray}
  WYS$^2$-SS (Ours) & \textbf{76.2} \tiny{$\pm$~0.2} & \textbf{80.7} \tiny{$\pm$~0.1} & \textbf{77.0} \tiny{$\pm$~0.1}\\
 \hline
\end{tabular}
\vspace{0.1cm}
    \end{minipage}
    \vspace{-0.8cm}
\end{table}

\subsection{Pseudo-labeling a large-scale unlabeled dataset}
\label{sec:pseudo}

Next, we use our best performing model WYS$^2$-SS %
to pseudo-label the large-scale \textit{unlabeled} HowTo100M dataset, introduced as $\mathcal{D}_U$ in Sec.~\ref{sec:data}.  Originally there are no temporal annotations in $\mathcal{D}_U$; we use our model to identify clips with %
visual narrations to obtain a larger $\mathcal{D}^\prime_{vn}$, retrain on it, and then test on VND Test and HTM-Align.

We compare to TAN~\cite{tan}, MIL-NCE \cite{miech19endtoend}---original model on which we designed the Temporal Alignment baseline
 but now pretrained on HowTo100M ---%
and CLIP \cite{clip}, a popular large-scale pretrained model. Table \ref{tab:vnd-and-align} (left two columns) shows the results.  Our method outperforms the baselines 
on both datasets. 
Since TAN is originally designed for alignment, we adapt it to our VND task by providing it short clips (instead of a long video) and checking for alignability; 
if it reports ``unalignable'', we take this to mean ``non-visual''.

\newremoved{Figure~\ref{fig:pseudo-finetune}(a) shows the results. At 0 pseudo-labels, we apply the model reported in Figure~\ref{fig:output-pr-curve} which is trained only with COIN and CrossTask.  Then, we successively increase the number of pseudo-labeled videos from $|\mathcal{D}_U|$ in steps of 25k up to 75k videos.\footnote{
While HowTo100M has 1.22M total videos, 75k represents the limits of our available  compute for training.}  \KGnote{update if we're going further} We randomly sample from those positively labeled by our model. 
We can observe a significant and monotonic increase in accuracy after adding the pseudo-labeled samples. Our model gains a full \textbf{7 points in mAP} yet uses no additional labels. In comparison, the baselines trained on labeled datasets (COIN and CrossTask) together with 75k videos perform significantly worse than our model. %
\KG{Notably}, %
using only 25k videos, our method \KA{performs similar to} %
the best baseline trained with 75k videos. %
}

This experiment illustrates that we can use %
manually labeled datasets ($\mathcal{D}_L$) like COIN and CrossTask to pseudo-label large-scale data ($\mathcal{D}_U$) like HowTo100M. %
Our framework allows us to enhance the power of large-scale unlabeled datasets without additional human labeling effort. For example, by pseudo-labeling 1.2M videos of HowTo100M, we increase the scale of the original annotation pool from COIN and CrossTask by 80,000$\times$, without any annotator effort. %
This idea to harness manually annotated datasets to obtain good quality video clips having visual narrations could be applied %
to remove non-visual training samples, thus saving training cost.

{
 \setlength{\tabcolsep}{1pt}
 \setlength{\extrarowheight}{1.5pt}
\begin{table}[t]
\captionsetup{font=footnotesize}
\begin{center}
\caption{Instructional video summarization accuracy %
on WikiHow test dataset using our features with the SoTA summarizer IV-Sum\cite{iv-sum}. A higher number implies better performance for all metrics.}
\label{tab:summarization}
\begin{tabular}{L{3.0cm}  C{3.7cm}  C{1.1cm}  C{1.1cm}  C{1.3cm}  C{1.3cm}}
\toprule
\multirow{2}{*}{Initialization} & \multirow{2}{*}{Method} & \multicolumn{2}{c}{F-Score} & $\tau$ \cite{kendall} & $\rho$ \cite{zwillinger1999crc} %
\\
\cline{3-4}  
& & Val %
& Test & Test & Test %
\\
\hline
MIL-NCE \cite{miech19endtoend} & CLIP-It w/ ASR \cite{clip-it} & 62.5 & 61.8 & 0.093& 0.191 \\
MIL-NCE \cite{miech19endtoend} & IV-Sum w/o ASR \cite{iv-sum} & 65.8 & 65.2 & 0.095& 0.202 \\
MIL-NCE \cite{miech19endtoend} & IV-Sum \cite{iv-sum} & 67.9 & 67.3 & 0.101 & 0.212 \\
  \rowcolor{Gray}
  WYS$^2$-SS (Ours) & IV-Sum \cite{iv-sum} & \textbf{69.0} \tiny{$\pm$~0.1} & \textbf{68.9} \tiny{$\pm$~0.1} & \textbf{0.106} \tiny{$\pm$~0.001} & \textbf{0.218} \tiny{$\pm$~0.002}
  \\ 
 \hline
\end{tabular}
\end{center}
\vspace{-0.5cm}
\end{table}

}

\vspace{-0.1cm}

\subsection{Alignability detection in long videos}

Next we apply WYS$^2$ for alignability detection, to allow further comparison with TAN~\cite{tan} in the authors' reported setting.   Recall that a video and narration are \emph{aligned} if the narration is relevant to what is shown, whereas it is a \emph{visual narration} if the visual content actually depicts the narrated action---a more selective criterion. Thus although our model is not trained for this task, it is applicable.
We provide our learned features as input to the TAN alignment head. Table \ref{tab:vnd-and-align} (right column) shows the results. %
We improve state of the art in alignability prediction. This result shows that our learned representation can be directly applied to alignability prediction. Note that the authors of TAN \cite{tan} updated the ROC-AUC to 75.1 following an error in paper's reported numbers (see \cite{tan-code}) and we compare against the corrected number.

\vspace{-0.1cm}

\subsection{Summarizing instructional videos}
\label{sec:summarization-results}

Having established the ability of WYS$^2$ to detect visual narrations and detect alignability, next we explore another downstream application: 
instructional video summarization, where the goal is to select which segments are essential  to understand the longer original video.
We augment the state-of-the-art instructional video summarizer IV-Sum \cite{iv-sum} with our features%
, and quantify performance using the authors' ground truth WikiHow video test set (see \cite{iv-sum} for details). 
Table \ref{tab:summarization} shows the results. We show F-score, Kendall's correlation $\tau$ \cite{kendall}, and Spearman's correlation coefficient \cite{zwillinger1999crc}---following \cite{iv-sum}. 
Replacing the MIL-NCE features used by IV-Sum with WYS$^2$ features, the output summaries are measurably better. %
We attribute this gain to the generalizability of our VND model. %
This shows the real-world applicability of our idea; such summaries are valuable to allow people to rapidly watch how-to's to learn new skills.  See Supp.~video for summary examples.

\audiocc{
\subsection{Visual narration detection from audio}

Finally, we present an interesting empirical discovery: %
 \emph{audio signals alone} contain relevance cues that indicate when a narration is about the demonstration taking place.

Table~\ref{tab:ablations}(right) shows the results. 
We see our simple audio classification framework outperforms the baselines.\footnote{Temporal Alignment and Overlapping Clips do not have explicit negative visual narration labels and hence cannot be used for this evaluation.} Our method is particularly better for typical recall values around 0.2--0.7.  This initial result suggests that the voice of the narrator itself lends clues to when the speech is about the action being performed---not what is being said, but how it is being said.  
Nonetheless, as expected, the absolute accuracy remains lower than when using both the visual and ASR-text inputs.
Figure~\ref{fig:audio-op} shows example audio predictions. 

}

\vspace{-0.2cm}

\section{Conclusion}

We propose \textit{What You Say is What You Show}, a novel approach for a new task to detect visual narrations for in-the-wild how-to videos. Using modest annotated data, we show how to automatically create labels for our task, and further boost training capabilities by pseudo-labeling large-scale unlabeled data. WYS$^2$ shows strong performance against competitive baselines and state-of-the-art video representations; %
it successfully improves the state-of-the-art in alignability prediction and instructional video summarization.
We additionally introduce a new %
benchmark that
will be released to the community. %

{\small
\bibliographystyle{ieee_fullname}
\bibliography{egbib}
}

\end{document}


\maketitle

\section{List of Contents}

This supplementary material includes the following additional details:

\begin{enumerate}
    \item A \textbf{video} showing 
    \begin{itemize}
        \item Positive and negative examples of automatically curated training set
        \item Positive and negative examples of mechanically annotated VND test set
        \item Additional examples of unseen HowTo100M video summarization including failure cases
        \item Examples of WikiHow instructional video summarization 
    \end{itemize}
    \item Details of manual annotation procedure adopted for creating the gold standard test set.
    \item Ablation study to explain the effects of changing the number of positive and negative clips, batch size, and learning rate.
    \item Limitations
\end{enumerate}

\section{Details of Manual Annotation Procedure}
\begin{figure*}
\centering
\includegraphics[width=\textwidth]{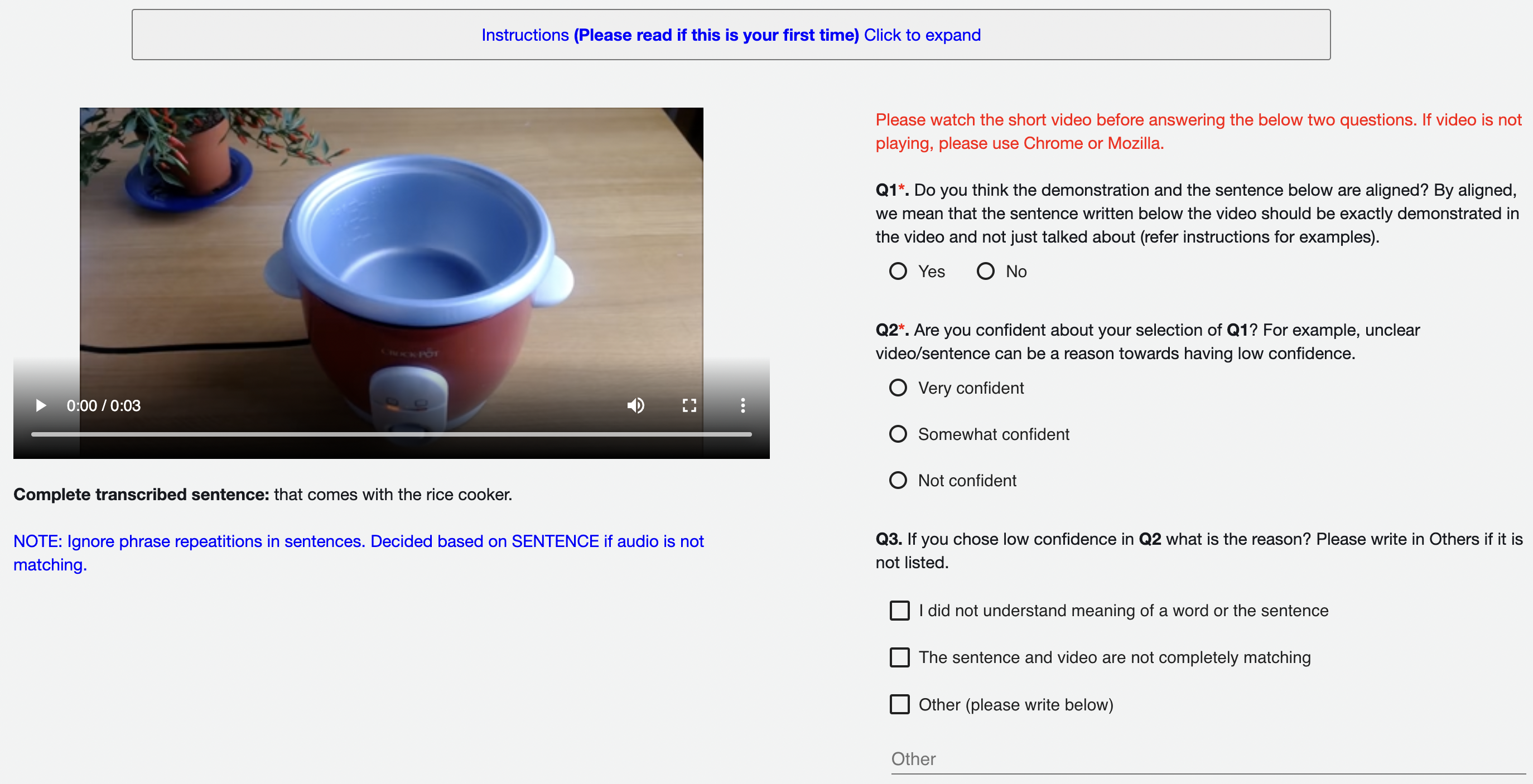}
\vspace*{-0.05in}
\caption{Screenshot of the AMT interface used to collect the gold standard test set for visual narrations.} %
\label{fig:interface}
\end{figure*}

We collected a gold standard VND test set (L201) to evaluate the performance of our method, to compare it against the baselines and SotA methods. We use Amazon Mechanical Turk (AMT) to collect human-annotated labels for visual narrations. We only allow human workers having a previous HIT acceptance rate of 98\% and passing a qualifying assessment consisting  of nine questions. The participants had to answer at least six out of the nine questions correctly to qualify.
For every instance, we first ask whether a video and a corresponding ASR text has a visual narration (only yes/no allowed). Next, we ask if they are confident about their choice with \textit{very confident}, \textit{somewhat confident} and \textit{not confident} as the options. Lastly, the third (not compulsory) questions asks for the reason of their low confidence. This question is aimed at determining whether a user failed to understand the meaning of some word or if the ASR text does not make sense. We also make sure the users watch the complete video clip before answering the questions.  Figure \ref{fig:interface} contains a screenshot of the interface visible to the human annotators. Also, the attached video \KA{and Fig 4 (main paper)} contains sample responses from the human annotators. \KA{The annotators were appropriately paid for their work through the same Amazon Mechanical Turk platform.}

\section{Ablations}
We select the values of the hyperparameters in our model based on performance on the validation set $\mathcal{D}_{vn}^{val}$ (disjoint from the gold standard test set used for all reported results). We present below ablations and analyses for these hyperparameters.

\textbf{Sentence Similarity Threshold ($c$).} To semantically match the key-step annotations and the narration, we use a Sentence Similarity model $\mathcal{S}(y, t)$ (L148). Positive samples are then chosen based on a threshold $c$. A higher $c$ implies a stricter requirement for semantic matching that results in a dataset with fewer positive samples. In contrast, a lower value of $c$ implies a larger dataset but allows $(y, t)$ pairs with lower semantic agreement to be labeled as positive.

\textbf{Number of negatives per positive sample ($|\mathcal{N}|$).} In our proposed method WYS$^2$-SS, we use an NCE objective function. We choose multiple negative samples for every positive video-text pair. We observe that choosing an excessively large number of negatives is detrimental. Instead, training with a smaller number of negative clips increases the validation performance.

{
 \setlength{\tabcolsep}{1pt}
 \setlength{\extrarowheight}{1.5pt}
\begin{table}[t]\footnotesize
\begin{center}
\begin{tabular}{L{1.1cm}| C{1.1cm} | C{1.1cm} | C{1.1cm} | C{1.1cm}}
&  $|\mathcal{N}|=3$  & $|\mathcal{N}|=5$ & $|\mathcal{N}|=7$ & $|\mathcal{N}|=9$ \\
\hline
  $c=0.4$& 67.4 & 68.0 & 67.4 & 68.0\\
  $c=0.5$& 67.2 & \cellcolor{gray!25}\textbf{68.3} & 67.1 & 67.2\\
  $c=0.6$& 66.2 & 66.3 & 66.5 & 65.0\\
 \hline
\end{tabular}
\end{center}
\caption{Ablation results on the choice of $|\mathcal{N}|$ and $c$  on the validation set $\mathcal{D}^{val}_{vn}$. 
}
\label{tab:ablation}
\end{table}
}

Table \ref{tab:ablation} shows the results of WYS$^2$-SS on the validation set for various choice of $|\mathcal{N}|$ and $c$. We choose the parameters that give the best result on the validation set.

\section{Limitations}

We propose the task of \emph{visual narration detection} and show that we can leverage keystep annotations to curate a dataset for VND training. Our training dataset curation uses sentence similarity between keysteps and ASR sentences which is prone to errors, as shown in Fig. 5 (main paper, last row). Despite this limitation, we show that our automatically curated train set improves over the baseline and psuedo-labeling the model on full-scale HowTo100M further boosts the performance.

\remove{
\subsection{Sentence similarity threshold}
To semantically match the key-step annotations and the narration, we use a Sentence Similarity model $\mathcal{S}(y, t)$ (L305). Positive samples are then chosen based on a threshold $c$. A higher $c$ implies a stricter requirement for semantic matching that results in a dataset with fewer positive samples. In contrast, a lower value of $c$ implies a larger dataset but allows $(y, t)$ pairs with lower semantic agreement to be labeled as positive. Figure \ref{fig:ablation-outputs}(a) shows the validation performance of our method WYS$^2$-SS for various values of $c$. The best performance is obtained at $c=0.5$, which was used for all experiments in the main paper.

\subsection{Number of negative pairs per positive sample}
In our proposed method WYS$^2$-SS, we use an NCE objective function. We choose multiple negative samples for every positive video-text pair. We observe that choosing an excessively large number of negatives is detrimental. Instead, training with a smaller number of negative clips increases the validation performance. Figure \ref{fig:ablation-outputs}(b) shows the performance on the number of negatives per positive used in the training.

\subsection{Number of positive pairs per training batch}
The Temporal Alignment baseline  \KG{modeled after~\cite{miech19endtoend}} uses multiple adjacent positive narrations per video clip. The choice of the number of positive narrations $|\mathcal{P}|$ is also chosen based on the validation performance. A higher value of $|\mathcal{P}|$ renders the approach robust to longer temporal misalignments but it also increases the possibility of different narration being labelled positive. Figure \ref{fig:num_pos} shows the performance as the number of positives is varied per training sample. All the other samples in the batch contribute as a negative sample for training, as in \cite{miech19endtoend}.

\begin{figure}[t]
\centering
\includegraphics[width=0.45\textwidth]{CVPR 2023/supp-figures/ablation_num_pos.pdf}
\caption{Ablations on the choice of positive pairs for Temporal Alignment \cite{miech19endtoend} baseline.
}
\label{fig:num_pos}
\end{figure}

\subsection{Batch size}
Figure \ref{fig:ablation-outputs}(c) shows a comparison of performances obtained with different batch sizes. We observe a better performance for a higher batch size. We choose a batch size of 128 for our experiments in the main paper.

\subsection{Learning rate}
Figure \ref{fig:ablation-outputs}(d) details our experiment with various learning rates. A learning rate of $10^{-5}$ yields the best validation performance.

}